# Submission Template for IET Research Journal Papers

# DuckSegmentation: A segmentation model based on the AnYue Hemp Duck Dataset

Ling Feng [1*], Tianyu Xie [1], Wei Ma [2], Ruijie Fu [1], Yingxiao Zhang [1], Jun Li [1], Bei Zhou [1,3*]

[1] College of Information Engineering, Sichuan Agricultural University, No. 46 Xinkang Road, Ya'an, 625000, Sichuan Province, China
[2] College of Mechanical and Electrical Engineering, Sichuan Agricultural University, No. 46 Xinkang Road, Ya'an, 625000, Sichuan Province, China.
[3] Ya'an Digital Agricultural Engineering Technology Research Center, No. 46 Xinkang Road, Ya'an, 625000, Sichuan Province, China.
[*] 202105857@stu.sicau.edu.cn; 12801@sicau.edu.cn;

**Abstract:** The modernization of smart farming is a way to improve agricultural production efficiency, and improve the agricultural production environment. Although many large models have achieved high accuracy in the task of object recognition and segmentation, they cannot really be put into use in the farming industry due to their own poor interpretability and limitations in computational volume. In this paper, we built AnYue Shelduck Dateset, which contains a total of 1951 Shelduck datasets, and performed target detection and segmentation annotation with the help of professional annotators. Based on AnYue ShelduckDateset, this paper describes DuckProcessing, an efficient and powerful module for duck identification based on real shelduckfarms. First of all, using the YOLOv8 module designed to divide the mahjong between them, Precision reached 98.10%, Recall reached 96.53% and $F1_{score}$ reached 0.95 on the test set. Again using the DuckSegmentation segmentation model, DuckSegmentation reached 96.43% mIoU. Finally, the excellent DuckSegmentation was used as the teacher model, and through knowledge distillation, Deeplabv3 r50 was used as the student model, and the final student model achieved 94.49% mIoU on the test set. The method provides a new way of thinking in practical sisal duck smart farming.
.

## 1. Introduction

With the continuous development of modern society and economy, the global consumption level has been increasing. People's living standards continue to improve, people's demand for poultry meat, eggs and other poultry-related products continue to increase, livestock and poultry farming industry has a broad space for development. Such a large-scale demand for livestock and poultry products will inevitably lead to the expansion of the scale of the farming industry, the traditional way of relying only on manual management is gradually falling behind, and in the context of the tight supply of feed grain, the need for soil resources for breeding, and the scarcity of water resources, the farming industry needs to continuously improve the quality and efficiency of production. Inefficient farming practices can increasingly exacerbate farming pollution, leading to an increased environmental burden and a departure from the concept of environmental protection [1].

The sparrow-feathered duck, commonly known as the "shelduck", is the main species of domestic duck and one of the most numerous, widely distributed and diverse species of domestic duck in the world. Duck breeding accounts for about 70% or more of the total waterfowl breeding and can be broadly divided into three types: meat, eggs, meat and eggs with high economic value. Large-scale shelduck farming can meet the huge market demand for poultry meat and eggs, but at the same time, it also faces pressure and challenges in many aspects[1][2]. As countries have discovered the importance of protecting the ecological environment and have made corresponding protection and restriction measures, resulting in less and less contraction of the poultry farming environment. Today, poultry farming has gradually developed towards green ecology. The current sisal duck breeding base needs to observe the duck situation in real time, count the ducks, analyze and summarize the suitable breeding mode according to the breeding environment so as to improve the duck production and quality. However, the current breeding base has the following problems: 1. workers do not always stay in the poultry house to monitor the duck flock 2. workers directly to count work, not only time-consuming and labor-intensive, and easy to make mistakes, counting errors 3. ducks move, gather and other behaviors further increase the difficulty of counting 4. workers are not able to accurately analyze the duck house environment, and choose the appropriate breeding method.

With the development of technology, many new methods of identification and monitoring have emerged in poultry farming. For example, Huishan Lu et al[3] designed a wireless sensor network-based body temperature monitoring system for real-time monitoring of poultry body temperature, which can quickly detect sick and dead poultry individuals. However, wearable devices do not have an effective battery life guarantee. Vroegindeweij et al [4] proposed a simple pixel-based classification method based on spectral reflectance properties for identifying and detecting eggs, hens, house elements and garbage using robots. Although the method has an accuracy of 80%, it is difficult to deploy in practice, and the accuracy is difficult to meet the practical application standards[5]. In recent years, due to the rapid development in the field of computer and artificial intelligence, research on the application of artificial intelligence technology to various other fields has been increasing, and the application of computer vision in agriculture is becoming more and more widespread, while the application of this technology to shelduckmonitoring is gradually making new progress.

For example, Girshick et al. proposed a pioneering CNN-based two-level object detection network, RCNN, in 2014 [6]. Such networks have excellent target detection accuracy and have made great progress in the field of computer vision, but their drawback is also obvious: redundant feature computation for a large number of overlapping proposals (more than 2000 frames from a single image) leads to extremely slow detection speed (14s per image using a GPU). Later in the same year, SPPNet[7] was proposed and overcame this problem. SPPNet effectively improves detection speed, but still has some drawbacks: first, training remains multi-stage, and second, SPPNet fine-tunes only its fully connected layers, while ignoring all previous layers[8]. They are difficult to deploy as well as use in sisal duck farms because they still suffer from high computational cost due to redundant operations. Lin et al[9] proposed a single-stage detector RetinaNet.RetinaNet greatly improves the robustness and generalization performance of the network using Feature Pyramid Network (FPN) as well as focal loss. However, because the network structure is more complex and requires more computational resources, and the performance of RetinaNet is weak for small targets and unstable when monitoring sisal ducks, the network still does not meet the requirements of both performance and accuracy in large-scale sisal duck breeding. Our idea is to use a single-stage powerful target detection algorithm, and the YoLoV8 algorithm is introduced in this paper. YoLoV8 has shown extremely high accuracy in practical farm applications, and due to the poor environment of the duck house, the algorithm effectively shows the ideal performance and accuracy to meet the needs of domestic sisal duck smart farming.

Image semantic segmentation is the most basic algorithm in the field of computer vision, and the good or bad image semantic segmentation determinesthe strength or weakness of subsequent algorithms for classification or recognition. Therefore, the implementation and application of an effective image semantic segmentation algorithm is of great practical importance. The UNet proposed by Ronneberger et al [10] has enabled a significant reduction in the amount of data for deep learning neural networks that previously required large amounts of annotated data for training, and has pioneered the application of neural networks for image segmentation. R Ranftl et al. proposed DPT [11] which when applied to semantic segmentation sets a whole new level of technology with 49.02% mIoU for ADE20K, but the DPT model can only perform well on specific tasks and may not perform well for others. Wang J et al. proposed OCRNet [12] a new kind of object context information - explicitly augmenting the contribution of pixels from the same class of objects when constructing the context information, computing the relationship between each pixel and each object region, and increasing the representation of each pixel with the object context representation. Liang-Chieh Chen et al. proposed the model DeepLabv 3+[13] to extend previous models by adding a simple but effective decoder module. The work is based on the spatial pyramid pooling technique and constructs a core network architecture by fusing the advantages of multiple models to obtain a well-performing, deep network structure-based codec without complex preprocessing, which further improves the accuracy and performance of the model. Yu C, Wang J et al. proposed BiseNetV1 [14] proposed a new approach to decouple the functions provided by the spatial information preservation and reception domains into two paths. A bilateral segmentation network (BiseNet) with spatial path (SP) and contextual path (CP) is also proposed. to encode more spatial information and expand the perceptual field while BiseNet V2 [15] proposed an efficient dual-path architecture on top of this, called bilateral segmentation network, to handle these spatial details and categorical semantics separately for highly accurate and efficient real-time semantic segmentation. A unified, simple and effective framework K-Net [16] was proposed by ShangTang Nanyang Technological University. It also outperformed all single models on the semantic segmentation task and achieved 54.3% mIoU on ADE20K. K-Net showed extremely high accuracy in real farm applications, and the algorithm effectively demonstrated the ideal performance and accuracy to meet the needs of domestic sisal duck smart farming.

In this case above, to better deploy neural networks in farms, we thought of knowledge distillation to solve such problems (KD)[17][18]. For deep neural networks [18], knowledge distillation refers to the technique of using a welltrained network to guide the training of another network, or using a large teacher model that works well to guide a lighter model for effective training, which results in improved detection accuracy. For the small size of the mahjong data set, so we used CWD channel knowledge distillation for the intensive prediction task by normalizing the activation maps in each channel. We then minimized the asymmetric Kullback-Leibler (KL) scatter of the normalized channel activation maps-which were transformed into a distribution for each channel-between the teacher and student networks. The activation of each channel tends to encode the salience of the scene category. For each channel, the student network is guided to pay more attention to mimic regions with significant activation values, leading to more accurate localization in intensive prediction tasks. For example, in object detection, the student network pays more attention to learning the activation of foreground objects [19]. Through knowledge distillation, it is evident that our network model has improved significantly in accuracy and is easier to deploy within real farms.

The paper is divided into six sections. The first part is a brief introduction to duck testing and duck splitting, among other things. The second part is for the description of the self-established AnYue Shelduck dataset. In the third part, the model proposed in the paper is analyzed and presented The fourth part is the model experimental design and presentation of results. Part V discusses. Part VI summarizes the paper with a description of the contributions.

## 2. Dataset

The dataset used for the experiments was taken at a shelduck farm in AnYue, Sichuan Province, Chian and we therefore refer to the dataset as the AnYue Shelduck dataset. The dataset was obtained using a camera Nikon D3500, the vertical and horizontal resolution of the images were 96 dpi, the dataset was shot with 60 sets of shelduck videos, each with a frame height of 1920 pixels and a frame width of 1080 pixels, yielding 5,213 shelduck images of varying quality levels. To ensure the quality and accuracy of the data, a total of 1951 high quality and valid shelduck images were obtained through manual data screening, and all images were sized at 853 pixels in width and 480 pixels in height. The shooting methods include long-distance shooting and close-up shooting, and the shooting angles include top-down shooting and parallel shooting. Top-down and long-distance shooting take up most of the dataset because this kind of shooting is more practical as it is more in line with the angle range of the system monitoring camera.

In addition, images containing only one duck are further extracted from the detection results. The distribution of the data is shown in Figure 1. x and y are the horizontal and vertical coordinates of the duck located in the image, and width and height are the length and width of the duck in the image. The distribution of duck locations generally follows a normal distribution, and the distribution of the dataset is closer to the natural situation.

The researchers labeled the annotations for target detection and semantic segmentation for each duck in each image. The marking follows the following principles:
(1) The category should be clear: for target detection, the category is set to Shelduck, and for subcategory detection, the category is set to Ruddy Shelduck.
(2) In the case where the target individual is occluded, truncated or blurred, the bounding box should be clearly defined: when the target individual is occluded or truncated, the bounding box should contain the key features of the target individual but not other individuals, and when the target individual is blurred, the samples are still involved in the training to improve the robustness of the model.
(3) Boundary checking after annotation: it is necessary to ensure that the bounding box coordinates are not on the image boundaries to prevent out-of-bounds errors during data enhancement.

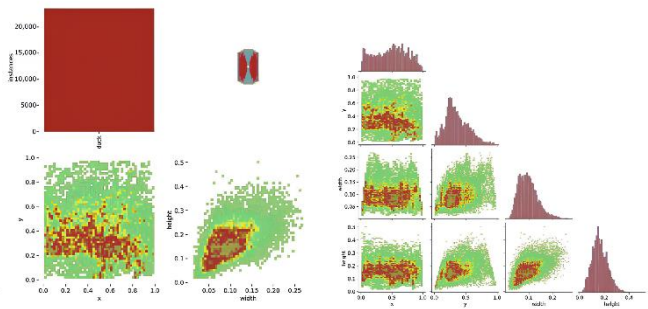

**Figure 1. Distribution of ducks in the dataset image. x and y are the horizontal and vertical coordinates of the ducks located in the image. Width and height are the length and width of the ducks in the image. The distribution of duck locations generally follows a normal distribution.**

Finally, before the dataset is fed into the model, in order to enhance the robustness of the model, we take different data preprocessing approaches for the training set and the test set, and we perform several data preprocessing for the training set:
(1) For the dataset, 10% of the shelduckimages are randomly selected for sharpening.
(2) Unify the size of each picture to 2048×512.
(3) Random crop size and random flip, the images fed into the network are randomly cropped to 512×512 size images, and there is a 50% probability that the test set will flip along the horizontal direction, the test data set is not used, for the final feed into the neural network.

## 3. DuckProcessing

A total of three tasks are performed for the AnYue shelduckdataset. The first stage of the task uses a target detection algorithm model. YOlOv8 [20] as the latest work of the yolo series, is a single-stage target detection algorithm. We feed the AnYue shelduckdataset into the improved YOLOv8 model and feed the output results into the segmentation model. The second stage of the task uses a segmentation model, in which the segmentation model K-Net is a uniform, simple and effective framework [16], while K-Net for content with very similar texture appearance, K-Net sometimes has difficulty to distinguish them from each other [16]. We also found the same difficulty in SAM [21] model to handle the AnYue sabal duck dataset, so we fed the sabal duck images that were accurately localized after passing the target detection task back into the K-Net model, using the target detection task to help the K-Net model target localization and enhance the model's ability to extract detailed information about the sabal duck. We update the backbone of K-Net, and in the decode head part of the model, we replace the standard convolution with fractionallystrided convolutions [22] product, and add Dropout layer [23] to improve the generalization ability of the model, while using *Lovász loss* [24]. In addition, in the third stage of the final task, in real embedded devices, it is not possible to carry a model with a large number of parameters, so for this reason knowledge distillation (CWD), using the model, allows the model to achieve a better balance between accuracy and efficiency. Figure 2 illustrates the general architecture of our approach.

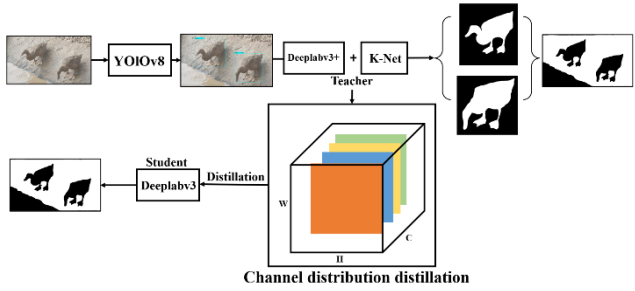

**Figure 2. AnYue DuckProcessing Overview.** First DuckProcessing facilitates sheldcuk instance segmentation through sheldduck detection, and imparts knowledge to smaller models through knowledge distillation.

This is an end-to-end model training. The information flow interaction process, first, the image passes through YOLOv8 and is used to extract features. These feature maps contain high-level abstract information about the image, such as texture, shape, and contextual information. The bounding box and the corresponding object class probabilities are generated afterwards. This information is passed to the segmentation module for subsequent processing. To realize the segmentation prediction, the segmentation module uses the studied DuckSegmentation, which outputs pixel-level predictions. During the information flow interaction, it is ensured that the two modules are spatially aligned, and the detection module obtains contextual information to help the segmentation module better understand the relationship between the object and its surroundings. Finally, the outputs of the detection and segmentation modules are typically synthesized to generate the final detection frame and object segmentation results.

A common problem in segmentation models is the difficulty in distinguishing content with similar texture appearance. This is because many segmentation models rely heavily on pixel-level features, such as color, texture, and shape, to distinguish between different objects. However, these models may encounter difficulties when two or more objects have similar textures or colors. In order to solve this problem, there are several possible solutions: using more complex features: in addition to the basic pixel-level features, more complex features can be used, such as contextual information, spatial relations and shape information of the object. For example, while the textures of a beach and a desert may be similar, their contextual information (e.g., the surrounding environment) may differ. By taking these additional features into account, the model may be able to better distinguish between objects with similar textures. Data augmentation: By augmenting the training data with augmentations such as rotation, scaling, clipping, and color transformations, the robustness of the model can be increased so that it can handle a wide variety of situations. This may help the model to better distinguish objects with similar textures. Use of multimodal data: In addition to using single image data, consider using multimodal data such as depth information, infrared information, or radar information. This additional data may provide additional information that helps to distinguish objects with similar textures. We hope to improve this situation with the proposed DuckProcessing model.

Finally, through experimentation and design, the best DuckProcessing model is to utilize YOLOv8 instead of the detection phase. Add Kernel Update Head to Deeplabv3+. At the same time Lovász loss replaces the old loss function in it, and utilizes fractionally-strided convolutions to replace the old convolutional kernel.

### 3.1. DuckDetectiont

YOLO (You Only Look Once) is a family of target detection algorithms whose basic idea is to feed the entire image into a deep neural network and then output a bounding bo x containing information about the location and category of all objects in the image. Unlike other target detection algorithms, YOLO enables real-time target detection and excels in terms of accuracy and speed.

YOLOv8 is a SOTA (State-of-the-Art) model, YOLOv8 replaces the C3 structure of YOLOv5 [25] with a C2F structure that has higher performance and better generalization capability, allowing for finer tuning of the model structure as shown in Figure 3. Secondly, the Head structure of the model is replaced by the excellent decoupled head structure, which separates the classification and detection heads, and the Anchor-Based [26] is abandoned and replaced by the more efficient Anchor-Free [26] structure, while YOLOv8 adopts the TaskAlignedAssigner [27] positive sample assignment strategy, which selects positive samples based on the weighted scores of classification and regression. The positive samples are selected and the training accuracy of the model is improved by using the Distribution Focal regression loss function [28] and the CIoU regression loss function [29], and still using the BCE classification loss function. Finally, the operation of Mosiac enhancement [30] is turned off in the last 10 epoch of the YOLOv8 training strategy, which successfully improves the accuracy. YOLOv8 network architecture is shown in Figure 4.

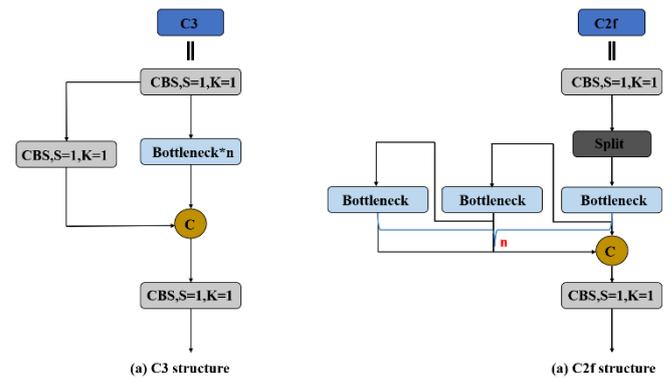

**Figure 3. Comparison of C3 and C2f Architectures.**

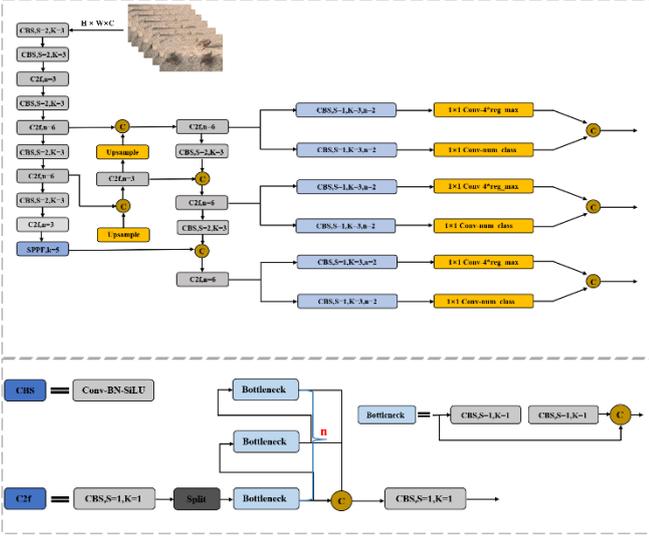

**Figure 4.** YOLOv8 Network Structure.

### 3.2. DuckSegmentation
#### 3.2.1 K-Net

In the process of smart shelduck farming management, semantic segmentation, instance segmentation and panoramic segmentation are required in different farming scenarios, so we need a unified, simple and effective framework to help us solve the shelduck farming management challenges.

K-Net is a simple, efficient, feature-rich and powerful framework, K-Net proposes a Kernel Update Head based on mask and feature map to dynamize the kernel, while K-Net achieves good mIoU values with faster inference on COCO and ADE20K datasets. Therefore, we decided to innovate on the basis of this framework and use it for the shelduck information extraction task.

K-Net uses Kernel Update Head to segment the image, which consists of group feature assembling, adaptive kernel update, and kernel interaction. The kernel update head first fuses the features of each group by group feature assembling, and then the group-based convolutional kernel has the initial perceptual capability, and then the adaptive kernel update obtains the convolutional kernel feature map obtained in the previous step to update theconvolution, so as to improve the representation capability of the convolutional kernel. Finally, the convolution obtained from the previous update is further processed in the kernel interaction, first using multihead attention and then a feedforward neural network, allowing each convolutional kernel to perceive contextual information from other groups, which allows the convolutional kernel to implicitly model and exploit the dependencies between mapping groups to finally obtain a new convolutional kernel. The semantic convolutional kernels can also be enhanced by augmenting the instance convolutional kernels with Kernel Update Head. the structure of Kernel Update Head is shown in Figure 5.

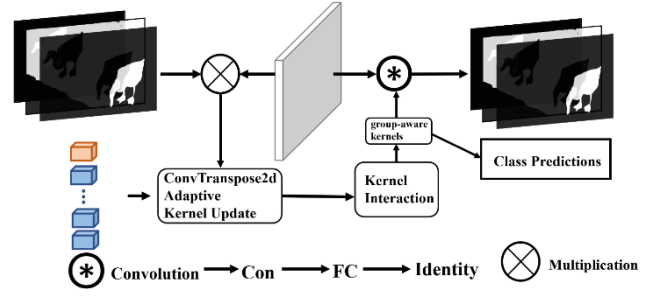

**Figure 5.** Kernel Update Head Structure. Next the Kernel Update Head was modified to meet our needs

#### 3.2.2 The structure of the network

In the target-detected shelduck breeding images, there may be different quality pictures, and in the second phase of the task we want to add more flexibility and functionality to the segmentation model to achieve precise adjustment of the output feature maps. Therefore, we changed the convolution operation in Kernel Update Head to fractionally-strided convolutions operation and added Dropout layer at the same time. Due to the flexibility of K-Net itself, which does not depend on a specific model, K-Net can attach its Kernel Update Head to any segmentation model to perform semantic segmentation. We experimentally find out the network model structure for the AnYue shelduck dataset pairs: we add Kernel Update Head to Deeplabv3+ [31] network architecture and pre-train the model with ResNet101-D8. And to improve the model accuracy, we used Lov´asz loss function that The relationship between pixels is taken into account.

#### 3.2.3 fractionally-strided convolutions

Wanting the K-Net network to learn the best way to upsample and the segmentation model to add more flexibility and functionality, we intend to use fractionally-strided convolutions convolutions. This method will not use a pre-defined interpolation method that has parameters that can be learned. Transposed Convolution is often also called Deconvolution and Fractionally-strided Convolution in some literature. Transposing the convolution causes the convolutional layer to perform feature extraction in the encoder and then restore it to its original size in the decoding layer, classifying every pixel of the original image. As shown in the Figure 6, a convolution operation is a many-to-one (many-to-one) mapping relation. For the convolution operation the inverse operation will be a one-to-many (one-to-many) mapping relationship. In essence, the transposed convolution is not a convolution, but we can think of it as a convolution and use it as such. We achieve dimensional upsampling by inserting zeros between elements in the input matrix for complementary purposes, and then the same effect as transposed convolution can be produced by ordinary convolution operations.

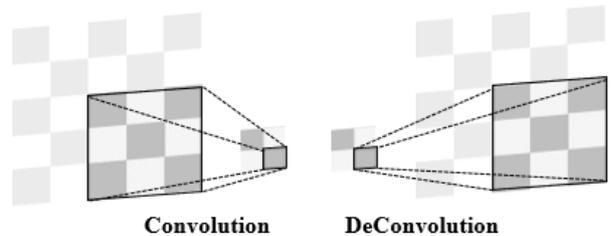

**Figure 6.** Schematic diagram of the structure of fractionally-strided convolution*s*

### 3.2.4 Lovász loss

In the AnYue shelduck dataset, the segmented shelduck objects sometimes occupy a part of the whole image area, CrossEntropyLoss insufficient to handle the highly imbalanced learning targets of masks. CrossEntropyLoss is insufficient to handle the highly imbalanced learning targets of masks and cannot consider the pixel-to-pixel relationship, while for Dice Loss [32] when the two segmented regions overlap less, the gradient of Dice Loss becomes small, which leads to training difficulties. Secondly, Dice Loss is less sensitive to the small regions in the segmentation results because they tend to be ignored, while the large regions contribute more to the total loss.

The Jaccard index, also referred to as the intersection-overunion (IoU) score, is commonly employed in the evaluation of image segmentation results given its perceptual qualities and scale invariance [24]. Therefore, it is straightforward to optimize the Jaccard index as a loss function, and given a vector of truth labels $y^*$ and a vector of predicted labels $y'$ the Jaccard index of class c can then be defined as:

$$J_c(y^*, y') = \frac{|\{y^* = c\} \cap \{y' = c\}|}{|\{y^* = c\} \cup \{y' = c\}|}$$

which gives the ratio in [0, 1] of the intersection between the ground truth mask and the evaluated mask over their union, with the convention that 0/0 = 1. A corresponding loss function to be employed in empirical risk minimization is

$$\Delta_{J_c}(y^*, y') = 1 - J_c(y^*, y')$$

However, the definition of Equation.2 is discrete and cannot be directly derived, so it cannot be used directly as a loss function. To overcome this problem, the authors make smooth extensions so that it can be used as the loss function of the segmentation network. The authors find that a piecewise linear convex surrogate based on Lovász extended submodular set functions is a good alternative to Jaccard loss.

In real applications, Lov´asz loss takes into account the relationship between pixels. Specifically, it uses a method called Lovász extension, which transforms the classification problem into a ranking problem and computes the loss using convex packet theory in the ranking problem. This method enables Lovász loss to rank samples one by one, thus ensuring an accurate consideration of the relative importance between samples. In addition, since Lovász loss can handle multi-label classification problems, it performs well in image segmentation tasks with multiple labels. It can optimize multiple labels simultaneously under a single model, thus avoiding the tedious process of training multiple models.

### 3.3. Knowledge Distillation

Although our model is theoretically powerful, the large number of our model parameters requires high hardware resource requirements when deployed in practical scenarios such as the AnYue sesame duck farm. As shown in Figure 7, The knowledge distillation method proposed by predecessors such as Geoffrey Hinton, Oriol Vinyals, and Jeff Dean [33], which can sacrifice some of the accuracy of the model to reduce the size of the number of model parameters, can train the model effectively. Knowledge distillation is the transfer of knowledge of the learned features from the already trained large model to the small model. We call the large model the teacher model and the small model the student model. Eventually the student model is taught by the teacher model to achieve the purpose of compressing the number of parameters.

For the traditional knowledge distillation approach, losses are calculated for the outputs of the student model and the teacher model so that the student model keeps approaching the teacher model. However, this approach only requires the small model to learn the output features of the large model and does not learn the structural features of the large model. We need a more efficient, stable and flexible knowledge distillation method.

The traditional knowledge distillation learning loss calculation, given the student model prediction $z_s$ and the teacher model prediction $z_t$, and given the distillation temperature $T$, the distillation loss can then be defined as

$$q_i = \frac{\exp\left(\frac{z_s}{T}\right)}{\sum_j \exp\left(\frac{z_t}{T}\right)}$$

To obtain better small models, we use the Channel-wise Knowledge Distillation (CWD) model proposed by Changyong Shu, Yifan Liu et alp. The CWD model performs the dense prediction task by normalizing the activation map in each channel, directing the student model to pay more attention to modeling regions with significant activation values regions with significant activation values, thus achieving more accurate localization in the intensive prediction task. A CWD schematic is shown in Figure 8.

In order to better utilize the knowledge in each channel, the authors convert channel activation into a probability distribution and propose a novel channel distillation paradigm so as to guide students to learn knowledge and features from the teacher model in an orderly manner. Given a teacher network $T$ and a student network $S$, the activation graphs from $T$ and $S$ are $y^T$ and $y^S$, respectively, then the channeled knowledge distillation loss can be expressed as:

$$\varphi(\phi(y^T), \phi(y^S)) = \varphi(\phi(y_c^T), \phi(y_c^S))$$

$\varphi()$ simultaneously converts the activation values into probability distributions as follows:

$$\phi(y_c) = \frac{\exp\left(\frac{y_{c,i}}{T}\right)}{\sum_{i=1}^{W \cdot H} \exp\left(\frac{y_{c,i}}{T}\right)}$$

Where c = 1, 2..., C indexes the channel; and i indexes the spatial location of the channel. t is a hyperparameter (temperature)

If we use a larger T, the probabilities become softer, which means we focus on a wider region of space for each channel. By applying softmax normalization, we eliminate the effect of the magnitude scale between large and compact networks. This normalization is helpful for thr knowledge distillation. If there is a mismatch in the number of channels between teachers and students, the number of channels in the student network is upsampled using a 1×1 convolutional layer. $\varphi()$ assesses the difference between the channel distributions of teacher and student networks. We use KL scatter:

$$\varphi(y^T, y^S) = \frac{T^2}{C} \sum_{c=1}^{C} \sum_{i=1}^{W \cdot H} \phi(y_{c,i}^T) \cdot \log\left[\frac{\phi(y_{c,i}^T)}{\phi(y_{c,i}^S)}\right]$$

KL scatter is an asymmetric metric. From Equation (4), we can see that if $\phi(y_{c,i}^T)$ is large, $\phi(y_{c,i}^S)$ should be as large as $\phi(y_{c,i}^T)$ to minimize the KL scatter. Otherwise, if $\phi(y_{c,i}^T)$ is very small, the KL scatter is less concerned with minimizing $\phi(y_{c,i}^S)$. Thus, student networks tend to produce similar activation distributions in foreground salience, whereas activation in background regions corresponding to teacher networks will have a smaller effect on learning. This asymmetric property of KL facilitates KD learning for intensive prediction tasks.

In this piece of paper, we use a relational knowledge refinement approach. We used the model that was improved according to the requirements in the 3.2 segmentation task as the teacher's model, and found the best student model through experimental comparisons to reduce the number of model parameters as much as possible while ensuring the accuracy and segmentation effect. The top performer in the experiment was Deeplabv3 [34].

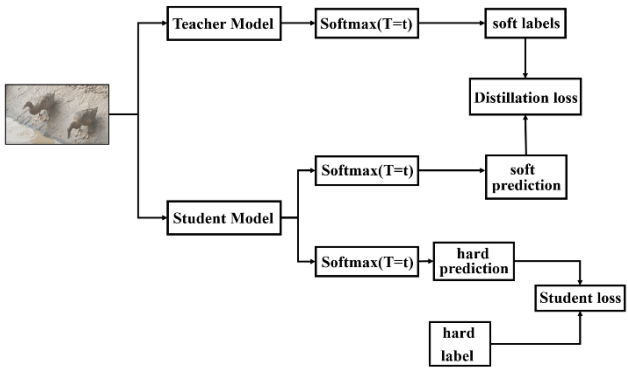

**Figure 7. Knowledge distillation model proposed by Hinton.**

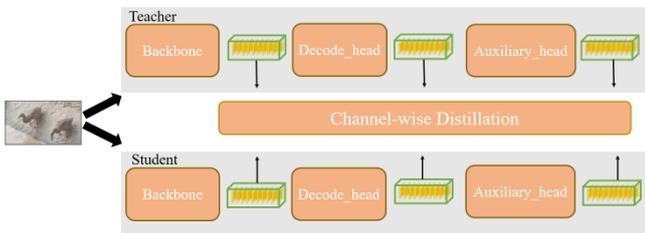

**Figure 8. The feature extraction extraction strategy of the Channel-wise Knowledge Distilla tion. Convenient student model learning more detailed information**

## 4. EXPERIMENTS AND RESULTS

In this subsection, we will perform data validation of the model, use experimental data to demonstrate the effectiveness of our proposed method, and visualize the recognition effect. In the target detection stage, compared with the traditional mainstream target detection models, the latest YOLOv8 model significantly outperformed other models in the information extraction task of the AnYue shelduck dataset, with a final accuracy of 98.1%, Recall of 96.53%, $F1_{score}$ of 0.95, $mAP@0.5$ of 0.977, $mAP@0.5: 0.95$ of 0.931 In the segmentation stage, the improved K-Net model algorithm reached 96.43 mIoU on the test set, which is 4.8 improvement over the original model, and in the knowledge distillation stage, the student model reached 94.49 mIoU on reduced computational complexity and parameter increase.

### 4.1. Evaluation indicators

To better evaluate the prediction quality and generalization ability of the model in this paper, we chose four evaluation metrics: IoU, Precision, Recall, $F1_{score}$, mAP, mDice, mIoU. Before calculating these metrics, several concepts need to be clarified. TP denotes the positive samples predicted by the model to be in the positive category. TN denotes the negative samples predicted by the model to be in the negative category. FP denotes the negative samples predicted by the model to be in the positive category. FN denotes the positive samples predicted by the model to be in the negative category.

The intersection over Union (IoU) is equal to the ratio of the intersection and the union between the "predicted border" and the "true border".The IoU of Prediction Box as B1 and Ground Truth Box as B2 is used as a metric to evaluate the correctness of the bounding box, i.e., the accuracy of the detection and thus the merit of the trained model.

$$\text{IoU} = \frac{B_1 \cap B_2}{B_1 \cup B_2}$$

Precision is the most commonly used classification performance metric. It can be used to express the accuracy of the model, i.e., the number of correct model identifications/total number of samples.The higher the accuracy of the model, the better the model is. It also reflects the ability of the model to distinguish negative samples, the higher the Precision, the stronger the ability of the model to distinguish negative samples. The calculation formula is as follows:

$$\text{Precision} = \frac{\text{TP}}{\text{TP} + \text{FP}}$$

Recall, also known as the accuracy rate, represents the ratio of the number of samples correctly identified by the model as positive classes to the total number of positive samples. A higher Recall indicates that more positive class samples are correctly predicted by the model. The calculation formula is as follows:

$$\text{Recall} = \frac{\text{TP}}{\text{TP} + \text{FN}}$$

The $F1_{score}$ is calculated based on the above accuracy and recall, and represents the reconciled average evaluation metric of PRECISION and RECALL. $F1_{score}$ is a combination of accuracy and recall, and a higher $F1_{score}$ indicates a more robust model. The calculation formula is as follows:

$$F1 = \left(\frac{2}{\text{Recall}^{-1} + \text{Precision}^{-1}}\right)$$
$$= 2 \cdot \frac{\text{Precision} \cdot \text{Recall}}{\text{Precision} + \text{Recall}}$$

AP (Average Precision): Average Precision, the average value of the highest Precision under different Recall (generally the respective AP is calculated separately for each category). The calculation formula is as follows:

$$AP = \sum_{k=1}^{N} P(k)\Delta r(k)$$

where N represents the number of all images in the test set, $P(k)$ represents the value of Precision when $k$ images can be recognized, and $r(k)$ represents the change of Recall value when the number of recognized images changes from $k − 1$ to $k$ (by adjusting the threshold).

mAP (mean Average Precision): the mean value of the average precision, the mean value of the AP for each category, is calculated as shown below:

$$mAP = \frac{\sum_{j=1}^{S} AP(j)}{S}$$

where $S$ is the number of all categories and the denominator is the sum of APs of all categories. The $mAP@0.5$ is the AP of all images in each category when the IoU is set to 0.5, and then the mAP is averaged over all categories. $mAP@0.5:0.95$ indicates the average mAP over a range of different IoU thresholds, from 0.5 to 0.95, in steps of 0.05.

The mDice coefficient is the average of the ratio of the intersection between the overall predicted and true edges of the sample to their sum, calculated as shown below:

$$mDice = \frac{\sum_{n=1}^{N} Dice(n)}{N} = \frac{1}{N}\sum_{i=1}^{N}\frac{2TP_i}{2TP_i + FP_i + FN_i}$$

The mIoU is a metric obtained by averaging the IoU of multiple objects and is one of the important metrics to measure the performance of the model. Higher values indicate more accurate detection results and better model performance. The calculation formula is shown below:

$$mIoU = \frac{1}{N}\sum_{i=1}^{N} IoU_i = \frac{1}{N}\sum_{i=1}^{N}\frac{TP_i}{TP_i + FP_i + FN_i}$$

where N is the number of objects, $IoU_i$ is the IoU value of the ith object, and $TP_i$, $FP_i$ and $FN_i$ denote the number of true positives, false positives and false negatives of the ith object, respectively.

Misclassification error (ME) is one of the ways to measure the performance of segmentation. It corresponds to the proportion of background pixels incorrectly assigned to the foreground. The value of ME is equal to 0 in the optimal case of thresholding and to 1 in the worst case.

$$ME = 1 - \frac{|B_O \cap B_T| + |F_O \cap F_T|}{|B_O| \cap |F_O|}$$

Where $B_O$ and $F_O$ denote the background and foreground of the ideal image respectively. $B_T$ and $F_T$ denote the background and foreground of the resultant image. The operator $|\cdot|$ denotes the number of elements in the set.

### 4.2. Experimental settings

All networks in this paper are written using the mainstream deep learning framework PyTorch. The hardware and software configurations of the devices are shown in Table 1

**Table 1. Hardware and software configuration table.**

| Configuration | Detail |
|---|---|
| CPU | 12th Gen Intel(R) Core(TM) i5-12400F |
| RAM | 32G |
| Graphics Card | NVIDIA GeForce RTX 3060 |
| Operating System | 64-bit Windows 10 |
| CUDA | 11.3 |
| Programming Language and Version | Python 3.7 |

The relevant parameters of the deep learning neural network model in this paper are shown in Table. 2,Table. 3. Due to the limitation of the experimental environment of the device, the batch size and the number of iterations of the deep learning model were set to 64 and 100, respectively, in the duck detection stage, while the conventional SGD was used for model optimization, and the initial learning rate was set to 0.01, as shown in Table 2. In the uck segmentation stage,the batch size and the number of iterations of the deep learning model in this paper are set to 2, respectively. This is the best performance. Additionally, Adamw is chosen for model optimization because of its high computational efficiency, faster convergence and good interpretability of the hyperparameters. The initial learning rate is set to 0.0001, as shown in Table 3. The training and test sets are divided in a ratio of 4:1 for all datasets used in this paper.

Training deep learning models usually requires a lot of computational resources In training AnYue Shelduck datesets, it takes an average of 90 seconds for every 100 rounds of epochs, and the final training of AnYue Shelduck datesets takes one hour and forty minutes. All images were taken at the Anyue poultry farm in Sichuan Province, China, and it took nearly two days to collect the data. Preparing the training data for the model requires a lot of labeling work, and we hired three professional data engineers who spent a total of three days to complete all the labeling. The cost of data storage and server is not counted.

**Table 2. Duck detection stage network model parameter list.**

| Parameters | Value |
|---|---|
| Batch Size | 64 |
| Epochs | 100 |
| Ratio of Training Set to Test Set | 4:1 |
| Initial Learning Rate | 0.01 |

**Table 3. Duck segmentation stage network model parameter list.**

| Parameters | Value |
|---|---|
| Batch Size | 2 |
| Epochs | 10000 |
| Ratio of Training Set to Test Set | 4:1 |
| Initial Learning Rate | 0.0001 |

### 4.3. Experiment
#### 4.3.1 DuckDetection

In Experiment 1, to verify the accuracy of the models, we compared the experimental effects of seven groups of models on the AnYue shelduck dataset, and we selected CenterNet [35], SSD [36], RatinaNet [37], YOLO X [38], YOLO v5, YOLO v7 [39], and YOLO v8. Four metrics, Precision, Recal, $F1_{score}$, $mAP@0.5$ and $mAP@0.5:0.95$, were used as accuracy metrics for shelduck detection. In the comparison of multiple groups of experiments, we obtained the most effective target detection model, YOLOv8, and the experimental data are shown in Table 4 and Table 5.

**Table 4. Performance of different models on the AnYue shelduck dataset**

| Method | Precision | Recall | $F1_{score}$ |
|---|---|---|---|
| CenterNet | 89.44% | 82.44% | 0.84 |
| SSD | 87.19% | 81.12% | 0.83 |
| RatinaNet | 86.96% | 79.75% | 0.81 |
| YOLO X | 89.64% | 80.30% | 0.82 |
| YOLO v5 | 91.62% | 85.41% | 0.88 |
| YOLO v7 | 92.90% | 87.32% | 0.90 |
| YOLO v8 | 98.10% | 96.53% | 0.95 |

**Table 5. Performance of different models on the AnYue shelduck dataset**

| Method | mAP@0.5 | mAP@0.5 : 0.95 |
|---|---|---|
| CenterNet | 0.796 | 0.630 |
| SSD | 0.782 | 0.660 |
| RatinaNet | 0.770 | 0.600 |
| YOLO X | 0.857 | 0.644 |
| YOLO v5 | 0.855 | 0.771 |
| YOLO v7 | 0.872 | 0.831 |
| YOLO v8 | 0.977 | 0.931 |

*4.3.2 DcukSegmentation*

For parameter fine-tuning, in order to find the optimal batchsize and learning rate combination, we experimented with multiple sets of batchsize and learning rate, as shown in Table 6. The model can perform best with this set of parameters. Eventually we found the optimal batchsize of 2 and learning rate of 0.0001. with this set of parameters, the DuckProcessing model gives maximum performance.

In some important experiments, three experiments were performed for each model, all adding the corresponding standard deviation.

**Table 6. Exploring optimal learning rate and batchsize on models**

| IR\BATH-SIZE | 2 | 3 |
|---|---|---|
| 0.01 | 38.71(±1.14) | 40.34(±1.02) |
| 0.001 | 91.35(±0.06) | 91.45(±0.21) |
| 0.0001 | 96.43(±0.24) | 94.35(±0.03) |

| IR\BATHSIZE | 2 | 3 |
|---|---|---|
| 0.01 | 0.5245(±0.014) | 0.5435(±0.011) |
| 0.001 | 0.9452(±0.004) | 0.9697(±0.003) |
| 0.0001 | 0.9614(±0.006) | 0.9364(±0.006) |

In Experiment 2, we obtained the most effective segmentation model DuckSegmentation. In order to verify the accuracy of the models, we first compared the experimental effects of eight groups of models on the AnYue shelduckdataset.

We chose K-Net, DPT, OCRNet, Deeplabv3plus, Deeplabv3, BisenetV2, BisenetV1, Unet. Two metrics, mDice, and mIoU, were used as accuracy metrics for sisal duck segmentation. Comparing in multiple groups of experiments, we obtained effective segmentation model K-Net, as shown in Table 7. We also plotted the scatter plots of mDice and mIoU based on the experimental results, as shown in Figure 9. The scatter plots can clearly reflect the performance of each model, and it can be seen from the scatter plots that the original K-Net has better generalization ability on our AnYue shelduck dataset.

**Table 7. Performance of different models on the AnYue shelduckduck dataset**

| Network | mDice | mIoU | ME |
|---|---|---|---|
| K-Net | 0.9563 | 91.63 | 0.0500 |
| DPT | 0.9399 | 88.67 | 0.0756 |
| OCRNet | 0.9501 | 90.51 | 0.0564 |
| Deeplabv3plus | 0.9432 | 89.26 | 0.0737 |
| Deeplabv3 | 0.9200 | 85.32 | 0.0520 |
| BisenetV2 | 0.9231 | 85.73 | 0.0812 |
| BisenetV1 | 0.9500 | 90.48 | 0.0499 |
| Unet | 0.8511 | 74.08 | 0.0302 |

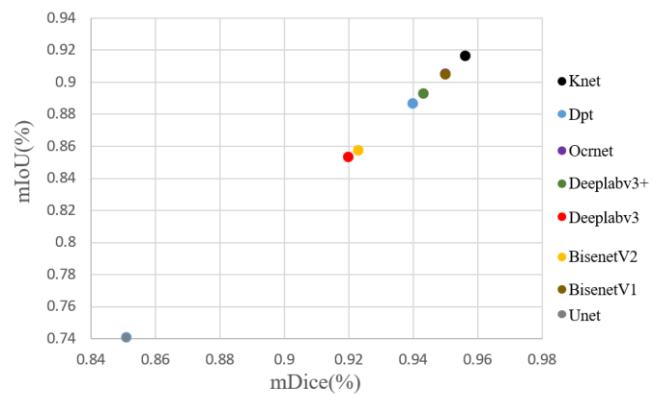

**Figure 9. Comparison of mDice and mIoU for different models. The initial K-Net model was**

In order to find the best K-Net model structure, we trained the model with five different methods based on K-Net instead, FCN [40], Deeplabv3, Deeplabv3plus, UperNet [41], and UperNet plus Swin Transformer [42]. The results are shown in the Table 8, comparing mIoU with mDice for each model and further improving the best model according to our needs.

**Table 8. Performance of different models on the AnYue shelduck dataset.**

| Network | mDice | mIoU | ME |
|---|---|---|---|
| K-Net_FCN | 0.9670 | 0.9363 | 0.0278 |
| K-Net_Deeplabv3 | 0.9281 | 0.8659 | 0.0648 |
| K-Net_Deeplabv3+ | 0.9678 | 0.9378 | 0.0431 |
| K-Net_Upernet | 0.8473 | 0.7351 | 0.1410 |
| K-Net_Upernet_Swint | 0.9195 | 0.8511 | 0.0606 |

We found that the model framework of K-Net (Deeplabv3+) performs best on our AnYue Shelduck dataset without considering the model size. For the K-Net (Deeplabv3+) model, further ablation experiments were conducted to verify the effectiveness of our improved model. We introduce fractionally-strided convolutions with Dropout in K-Net in Kernel Update Head, and also introduce different loss functions. The ablation results are shown in Table 9.

**Table 9. The model that combines Kernel Update Head with deeplabv3plus is called KUD. The KUD model is tried on the basis of whether to use fractionally-strided convolutions or not, the model that uses fractionally-strided convolutions is referred to as KUD-fs, and the one that does not is called KUD-Nfs. A second attempt was made to compare model strengths and weaknesses with or without Dropout; models that used Dropout were called KUD-D and those that did not were called KUD-ND.**

| Network | mIoU | ME | mDice |
|---|---|---|---|
| KUD-Nfs-ND | 93.78(±0.16) | 0.0410(±0.003) | 0.9678(±0.06) |
| KUD-Nfs-D | 93.62(±0.03) | 0.0444(±0.002) | 0.9669(±0032) |
| KUD-fs-ND | 95.04(±0.10) | 0.0440(±0.006) | 0.9691(±0.0011) |
| KUD-fs-D | 95.77(±0.11) | 0.0304(±0.003) | 0.9696(±0.0023) |

Different loss functions can have an impact on the convergence speed, robustness and handling of outliers of the model. It is usually necessary to choose the most appropriate loss function according to the characteristics of the specific task and dataset. We obtained the best performing KUD-fs-D model, on the basis of which we explored the impact of different loss functions on the results as shown in Table 10.

**Table 10. Based on the KUD-fs-D model, the effect of different loss functions on the model performance is explored.**

| | ME | mDice | mIoU |
|---|---|---|---|
| CrossEntropy Loss | 0.0242(±0.011) | 0.9713(±0.012) | 94.44(±1.13) |
| FocalLoss | 0.0250(±0.009) | 0.9703(±0.05) | 94.26(±0.62) |
| LovaszLoss | 0.0144(±0.01) | 0.9714(±0.09) | 96.43(±0.10) |
| DiceLoss | 0.0377(±0.08) | 0.9715(±0.01) | 95.63(±0.03) |

Finally, we obtained the state-of-the-art K-Net Deeplabv3plus+ fractionally-strided convolutions(Dropout)+ $Lov\acute{a}szLoss$ segmentation model, which we call DuckSegmentation model, on the AnYue shelduck dataset. To this end, we compared the improved model with the original model and trained the predicted training trajectory 10,000 times on the AnYue Shelduck dataset, as shown in Figure 10. It clearly reflects that the state-of-the-art model performs better on this dataset, with higher accuracy and faster convergence

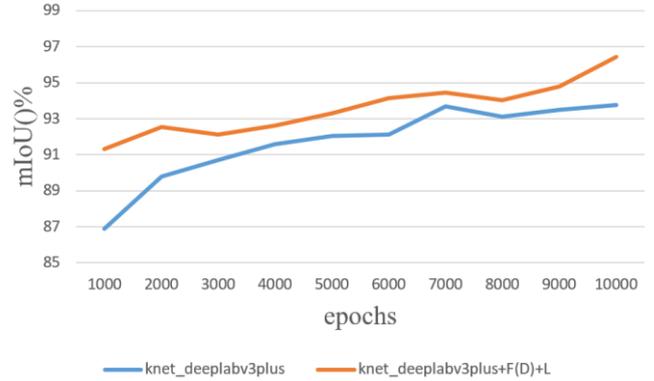

**Figure 10. Comparison of DuckSegformer and K-Net Deeplabv3plus training trajectories.**

Eventually the complete improved model becomes DuckProcessing and we compare DuckProcessing with other segmentation models. By comparing the metrics on the AnYue shelduck dataset, as shown in Table 11, to highlight the credibility of our model

**Table 11. Comparison of the final complete DuckProcessing model with the popular Mackerel Duck segmentation model.**

| Network | mDice | mIoU | ME |
|---|---|---|---|
| DuckProcessing | 0.9714(±0.023) | 96.43(±0.22) | 0.0144(±0.021) |
| DPT | 0.9399 | 88.67 | 0.0756 |
| OCRNet | 0.9501 | 90.51 | 0.0564 |
| Deeplabv3plus | 0.9432 | 89.26 | 0.0737 |
| Deeplabv3 | 0.9200 | 85.32 | 0.0520 |
| BisenetV2 | 0.9231 | 85.73 | 0.0812 |
| BisenetV1 | 0.9500 | 90.48 | 0.0499 |
| Unet | 0.8511 | 74.08 | 0.0302 |

*4.3.3 Knowledge Distillation*

In Experiment 3, we used the DuckSegmentation model obtained in Experiment 2 as the teacher model in the CWD knowl edge refinement process. Because our DuckSegmentation has a large number of parameters, Params: 75.78 M, we perform knowledge distillation on DuckSegmentation to enable efficient deployment, and select four models, Deeplabv3, PSPNet [43], FCN, and UperNet, as student models for feature learning. Through comparison, it was finally determined that Deeplabv3 had the best learning ability, and we compared the training trajectory of Deeplabv3 with the student model, as shown in Figure 11 and Table 12.(The backbone of the student models are all ResNet, writing as r.) The experiments show that using the improved segmentation model DuckSegmentation as the teacher model can teach the student model features well, while using

Deeplabv3 as the student model can learn the features of the teacher model well while reducing the size of the model.

Table 12. Performance of different models on the AnYue Shelduck dataset

| Student Models | mIoU | ME |
| --- | --- | --- |
| Deeplabv3 r50 | 94.49 | 0.0214 |
| Deeplabv3 r18 | 90.44 | 0.0442 |
| Deeplabv3+ r50 | 90.46 | 0.0464 |
| Pspnet r50 | 91.33 | 0.0337 |
| Fcn r50 | 89.77 | 0.1140 |
| Upernet r50 | 92.11 | 0.0325 |

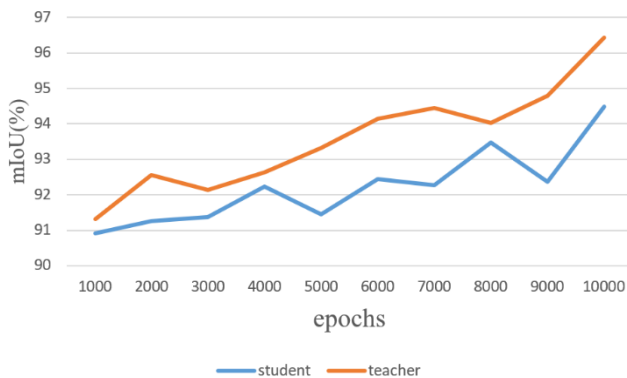

Figure 11. Comparison of the training trajectories of Deeplabv3 as a student model and Duck Segformer as a teacher mode

### 4.3.4 Effect show

As shown in Figure 12, We compare the processing of our DuckSegmentation, K-Net Deeplabv3plus and our original K-Net for the same set of AnYue shelduck images.

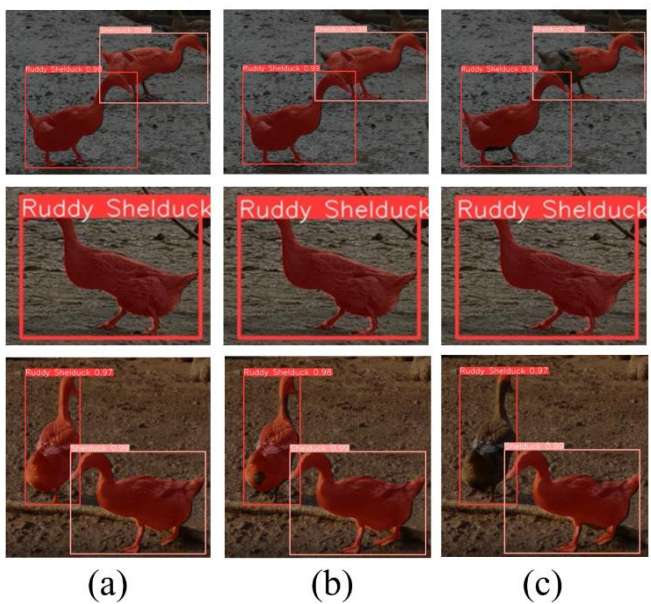

Figure 12. Surveillance on Shelduck and Ruddy Shelduck. Column (a) is the visualization of the segmentation effect of DuckSegformer proposed in this paper, column (b) is the visualization of the segmentation effect of K-Net Deeplabv3plus, and column (c) is the visualization of the segmentation effect through our original K-Net.

Eventually DuckPorcessing has been attempted on other datasets. SDSCNet [44] is a model that focuses on goose instance segmentation, and DuckPorcessing is a model that unifies semantic segmentation, instance segmentation, and panorama segmentation based on K-Net. Having been fortunate enough to contact the SDSCNet authors, as shown in Table 13. Comparison of detection accuracy results of goose instance segmentation network in SDSCNet, we compared DuckPorcessing before the introduction of the deeplabv3+ model with SDSCNet on the Goose dataset.

Table 13. Comparison of detection accuracy results of goose instance segmentation network in SDSCNet

| Model | mAP | mAP@0.5 | mAP@0.75 |
| --- | --- | --- | --- |
| SDSCNet | 0.602 | 0.933 | 0.710 |
| Mask R-CNN | 0.630 | 0.886 | 0.745 |
| YOLACT | 0.470 | 0.788 | 0.553 |
| SOLO | 0.599 | 0.892 | 0.724 |
| SOLOv2 | 0.587 | 0.884 | 0.782 |

Table 14. Performance of DuckProcessing on the Goose dataset. DuckProcessing1 has AnYue Shelduck Datesets training weights, while DuckProcessing2 has no training weights

| | mIoU | mDice | ME |
| --- | --- | --- | --- |
| DuckProcessing1 | 94.52 | 0.9712 | 0.302 |
| DuckProcessing2 | 89.69 | 0.9457 | 0.508 |

## 5. Discussion

With the rapid development in the field of computer vision, for example, like SAM and other large model feature extraction methods are rapidly emerging, but these large models often suffer from poor interpretation, large number of parameters and other problems, even after various ways to achieve real-time segmentation or deployment of models in real applications. Meanwhile more and more researchers are exploring how to reduce the computational complexity and the number of parameters of the model while ensuring the accuracy of the model. In the field of muggle duck recognition, there are also many researchers trying to use different approaches to solve this problem. For example, some researchers make more advanced network structure to implement shelduck recognition, Kailin Jiang proposed CBAM-YOLOv7 to implement shelduck recognition [1], and finally Precision reached 95.8%, Recall reached 93.64%, F1score reached 0.95, $mAP@0.5$ reached 97.57%, $mAP@0.5：0.95$ reached 65.50% [44]. However, these methods usually have a large number of parameters and computational complexity, which makes it difficult to efficiently implement shelduck identification under resource constraints.

Compared with previous work, the DuckSegmentation model proposed in this paper has better interpretability while ensuring model accuracy, and through knowledge distillation techniques, we also obtain a student model that can be practically applied and can be deployed and used in real-world scenarios.

Our DuckSegmentation model exhibits good performance in the field of muggle duck recognition. Compared with traditional deep learning-based methods, our model has higher recognition accuracy and less computational complexity, which makes our model more suitable for deployment in real-world applications. In addition, our model has better interpretability, which is very important for the staff of Anyue duck farms, as they need to understand how the model makes identification decisions for better management and maintenance.

In addition to the field of shelduck recognition, knowledge distillation techniques have been widely used in other fields as well. For exampleknowledge distillation techniques are also widely used in the fields of speech recognition, natural language processing, and image classification. By transferring the knowledge from the teacher model to the student model, the computational complexity and the number of parameters of the model can be effectively reduced and the generalization performance of the model can be improved. Therefore, knowledge distillation techniques have great potential for practical applications and can help us design more efficient and accurate models for different scenarios and tasks.

In the future, we hope to combine incremental learning with knowledge distillation. In the process of deploying different farms, because the data under many farms is constantly increasing, and at the same time new tasks will keep emerging, if the model is retrained every time, it will waste time and computational resources, and also lead to the degradation of model performance. We hope to save time and improve the efficiency of the model by incrementally learning the student model, retaining repetitive and useful information, and learning only new features and information brought by new data and new scenarios.

Overall, the combination of incremental learning and knowledge distillation is a very challenging problem that requires a combination of several factors such as model stability, accuracy, and generalization, and future research needs to further explore how to improve the efficiency and performance of learning.

For the DuckProcessing part, we can further explore how to further improve the performance and accuracy of the model for more complex and diverse real-world application scenarios. We can try to use more advanced network structures and algorithms, such as Transformer, Attention mechanism, etc., to further improve the performance and accuracy of the model. Also, we can explore how to combine the methods proposed in this paper with other techniques, such as migration learning and reinforcement learning, to further improve the performance and accuracy of the model.

In summary, this paper addresses the segmentation of the AnYue shelduck dataset identified by YOLOv8 target detection using the AnYue shelduck dataset as the research object. Using the unified, simple and effective K-Net model as the base segmentation architecture, the state-of-the-art DuckSegmentation model is obtained by continuously improving the model based on the K-Net model according to the actual requirements. Through ablation experiments, we verified the validity and advancedness of the model. Finally, the advanced model obtained was distilled for knowledge, and the student model was successfully obtained for practical application. The experimental results show that our advanced model most teachers can teach the student model well, so that our student model achieves a good recognition accuracy, computational complexity and number of parameters.

## 6. Conclusion AND Contribution

Overall, the main contributions of this study are as follows:

1. In this paper,annotating and processing the images of shelducks from AnYue farms in Sichuan, China, and establishing a dataset of AnYue shelducks.

2. In this paper,designing AnYue DuckProcessing module, which provides a new idea for intelligent shelduck farming.

3. In this paper,propose a high accuracy model for extracting information of shelducks– AnYue DuckSegmentation, which provides a new idea for the application of deep learning in the field of shelduck information extraction. This paper is divided into five parts.

In this paper, a dataset of Anyue duck farming with a distribution closer to the natural situation is established, and the DuckProcessing model is proposed. In the improved DuckProcessing recognition accuracy reaches 98.10% and MIOU for the segmentation task reaches 96.43%. In order to be able to deploy the model in line with the real situation, the improved model was subjected to knowledge distillation, and the generalization and feature extraction capabilities of the DuckProcessing model were imparted to the Deeplabv3 student model. Make Deeplabv3 student model has better performance in the information extraction task of An Yue Shelduck dataset, which is suitable for deploying the model in real applications.

In this paper, we established the AnYue shelduck dataset and proposed a DuckProcessing module in order to facilitate the management of local duck breeders. In the DuckProcessing module, the segmentation model is sometimes difficult to distinguish content with very similar texture appearance from each other. To facilitate the segmentation model , in the first stage, the information of the localized shelduck is transferred to the segmentation model through the YOLOv8 target detection stage. In the second stage, we introduce Deeplabv3plus based on the Kernel Update Head in K-Net In addition, in order to adapt the model to more farms and be more flexible and functionally new, we designed the DuckSegmentation model, using fractionally-strided convolutions and Dropout to improve the performance and stability of the model, increase the generalization ability of the model, and optimize and improve the mIoU through the Lov´asz loss function to finally obtain our state-of-the-art DuckSegmentation model. The mIoU of the segmentation task reached 0.9643 on our AnYue shelduck dataset. Finally, the state-of-the-art DuckSegmentation segmentation model is subjected to knowledge distillation, and the generalization ability and feature extraction ability of the DuckSegmentation model are imparted to the Deeplabv3 student model with a smaller number of parameters, so that the Deeplabv3 student model has a better performance in the shelduck breeding information extraction task and is suitable for the deployment of the model in practical applications. The Deeplabv3 student model achieves an mIoU of 0.9449 for the segmentation task on the AnYue Shelduck dataset. In discussing future directions for segmentation model improvement, proposing a more straightforward, unified, and lightweight segmentation model as an alternative to the large K-Net framework is a promising research direction. The following are some possible specific ideas and potential directions to help researchers and practitioners seeking efficient segmentation solutions: (1). Backbone Architecture: considering simplifying the structure of the backbone network to reduce unnecessary complexity. Techniques such as using lighter weight convolutional modules, reducing the number of layers, or employing depth-separable convolution to improve computational efficiency can be explored. (2). Self-supervised learning: developing new learning methods to reduce reliance on large amounts of labeled data. Self-supervised learning techniques can help models utilize limited labeled data more effectively. (3). Hardware acceleration: specialized hardware (e.g., GPUs, TPUs, etc.) is utilized to accelerate the inference process and improve the real-time performance of the segmentation model. (4). Model compression and pruning: investigating how to reduce the size of models and reduce memory and computational requirements through techniques such as pruning and quantization. These directions, help to develop more straightforward, unified and lightweight segmentation models. Researchers and practitioners can conduct in- depth studies in these areas to provide efficient segmentation solutions.These results provide a new option for the breeding of AnYue ducks, and also provide a valuable reference for research in the field of computer vision.

## 7. Declarations

• Funding

This project was funded by Sichuan Agricultural University Research Grant 202310626001x. National College Students' innovation and entrepreneurship training program(No. 202310626001X).

• Conflict of interest

All authors certify that they have no affiliations with or involvement in any organization or entity with any financial interest or nonfinancial interest in the subject matter or materials discussed in this manuscript.

• Ethics approval

The datasets generated during and analysed during the current study are not publicly available due to contains the AnYue farm business information but are available from the corresponding author on reasonable request.

• Consent for publication

Not applicable.

• Availability of data and materials

All data generated or analyzed during this study are included in this published article.

• Authors' contributions

All authors contributed to the study's conception and design. Material preparation, data collection and analysis were performed by Ling Feng and Tianyu Xie. The first draft of the manuscript was written by Ling Feng, and all authors commented on previous versions of the manuscript. All authors read and approved the final manuscript.

**Example References**